\def\arxivversion{1}
\newcommand{\arxivauthors}{%
  Tianshu Wu\textsuperscript{1},
  Tianxiang Gui\textsuperscript{2},
  Yue Chen\textsuperscript{1},
  Tingxu Liu\textsuperscript{1},
  Ziye Peng\textsuperscript{3}\\
  Guangjie Yang\textsuperscript{2},
  Xiangqi Kong\textsuperscript{3},
  Ye Shi\textsuperscript{2},
  Jingya Wang\textsuperscript{2},
  Hao Dong\textsuperscript{1}\\[0.5em]
  \textsuperscript{1}CFCS, School of Computer Science, Peking University\\
  \textsuperscript{2}ShanghaiTech University\quad
  \textsuperscript{3}Beihang University
}
\documentclass[10pt,twocolumn,letterpaper]{article}

\ifdefined\arxivversion
  \usepackage[pagenumbers]{cvpr}
\else
  \usepackage[review]{cvpr}
\fi

\def\confName{CVPR}
\def\confYear{2027}

\usepackage{amsmath,amsfonts,bm}

\def\eqref#1{equation~\ref{#1}}

\def\1{\bm{1}}

\DeclareMathAlphabet{\mathsfit}{\encodingdefault}{\sfdefault}{m}{sl}
\SetMathAlphabet{\mathsfit}{bold}{\encodingdefault}{\sfdefault}{bx}{n}

\usepackage{tikz}
\usetikzlibrary{arrows.meta}
\usepackage{tabularx}
\usepackage{longtable}
\usepackage{needspace}
\usepackage{float}
\usepackage{capt-of}
\usepackage{flafter}
\usepackage{stfloats}
\usepackage{cuted}
\definecolor{cvprblue}{rgb}{0.21,0.49,0.74}
\usepackage{xr-hyper}
\usepackage[pagebackref,breaklinks,colorlinks,allcolors=cvprblue]{hyperref}

\ifdefined\arxivversion
  \makeatletter
  \patchcmd{\@maketitle}{\vskip .375in}{\vskip 0pt}{}%
    {\PackageError{sugarv2}{Cannot adjust title top spacing}{Check cvpr.sty}}
  \patchcmd{\@maketitle}{\vspace*{12pt}}{\vspace*{0pt}}{}%
    {\PackageError{sugarv2}{Cannot adjust author bottom spacing}{Check cvpr.sty}}
  \patchcmd{\@maketitle}{\vspace*{24pt}}{\vspace*{6pt}}{}%
    {\PackageError{sugarv2}{Cannot adjust title-author spacing}{Check cvpr.sty}}
  \makeatother
\fi

\newcommand{\methodname}{SUGARv2}
\title{\methodname{}: Learning Diverse Visual Humanoid Loco-Manipulation Skills\\
from Synthesized Interaction References}
\ifdefined\arxivversion
  \author{\arxivauthors}
\else
  \author{Anonymous authors}
\fi

\begin{document}
\hypersetup{allcolors=black}
\raggedbottom
\maketitle
\begingroup
\ifdefined\arxivversion
\setlength{\stripsep}{8pt plus 2pt minus 2pt}
\fi
\begin{strip}
\par
\centering
\includegraphics[width=\textwidth]{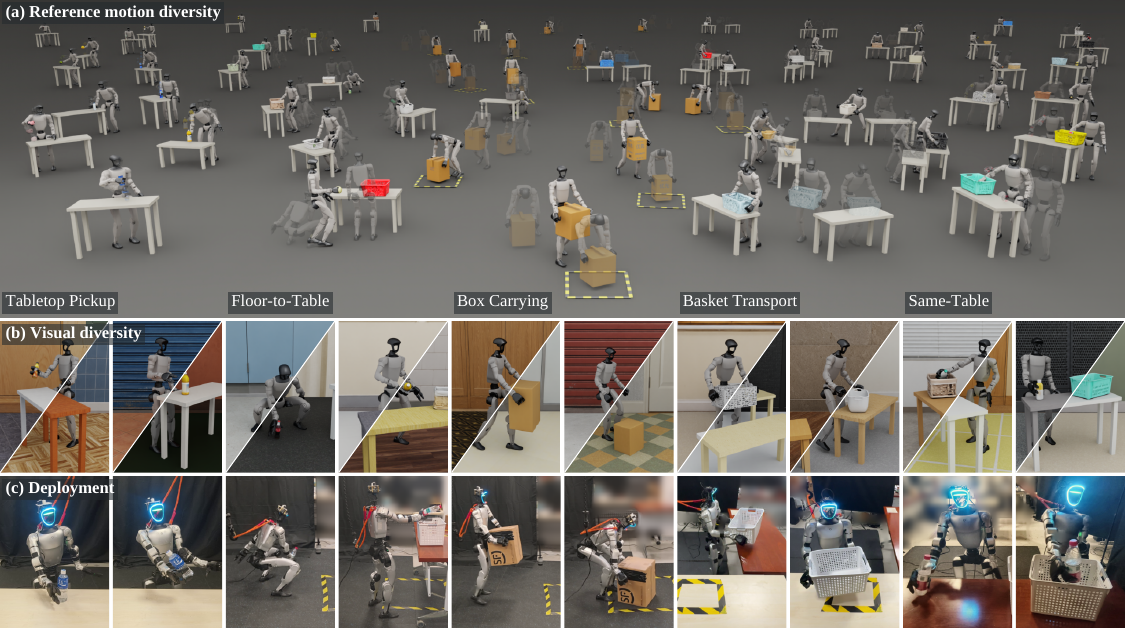}
\par
\captionsetup{hypcap=false}
\captionof{figure}{SUGARv2 combines interaction reference synthesis with visual policy learning across tasks. (a) Synthesized reference motions across five tasks and spatial configurations. (b) Visual diversity with varied object assets and scene appearances. (c) Real-robot deployment.}
\label{fig:teaser}
\captionsetup{hypcap=true}
\end{strip}
\endgroup

\begin{abstract}
\looseness=1
Scaling up visual humanoid loco-manipulation skill acquisition requires reducing the human effort needed as tasks and spatial configurations expand, particularly in demonstration collection and task-specific training design. We present SUGARv2, a unified framework for learning diverse visual humanoid loco-manipulation skills through interaction reference synthesis and a shared training pipeline. To obtain diverse references at low cost, we adapt reusable task templates to different spatial configurations, synthesizing coordinated robot and object motions. To reduce task-specific reward design and training effort, we use reference-based imitation as a common learning objective across tasks. We first train privileged teachers to execute these references, then train visual policies through teacher distillation and reinforcement learning with the same reference-based mimic rewards, using diverse object instances and visual domain randomization to support visual generalization and sim-to-real transfer. Across five humanoid loco-manipulation tasks, SUGARv2 achieves a mean success rate of 91.0\% in simulation. Deployment on a real humanoid robot further demonstrates successful sim-to-real transfer for whole-body manipulation. These results demonstrate the framework's potential for learning diverse visual humanoid loco-manipulation skills.
Project Page: \href{https://tianshuwu.github.io/sugarv2-humanoid/}{\textcolor{cvprblue}{\mbox{tianshuwu.github.io/sugarv2-humanoid/}}}.
\end{abstract}

\clearpage
\flushbottom
\section{Introduction}
\label{sec:introduction}
\suppressfloats[t]

Scaling up visual humanoid loco-manipulation skill acquisition requires reducing the human effort needed for each new task and spatial configuration. Existing approaches rely on manually collected interaction data or task-specific training designs, making it costly to extend skill learning across tasks and spatial configurations. HumanoidMimicGen~\citep{lin2026humanoidmimicgen} expands teleoperated demonstrations through whole-body planning, but depends on initial demonstrations and uses decoupled upper- and lower-body control. GRAIL~\citep{xie2026grail} avoids teleoperation and supports whole-body control, but requires video generation and motion reconstruction for every spatial configuration, incurring high costs and potentially producing physically implausible interactions. Task-specific policy learning methods~\citep{he2025viral,xue2025doorman,rayyan2026fetchman} acquire visual loco-manipulation skills through tailored training designs, requiring further engineering for new tasks. These dependencies make scaling skill acquisition costly in data preparation and training design (Table~\ref{tab:comparison}). Motivated by this, we present SUGARv2, a unified framework for learning diverse visual humanoid loco-manipulation skills through interaction reference synthesis and a shared training pipeline.

\newcommand{\cmark}{%
  \tikz[baseline=-0.55ex]\draw[green!52!black,line width=1.05pt,line cap=round,line join=round]
    (0,0.22ex)--(0.42ex,-0.22ex)--(1.22ex,0.78ex);%
}
\newcommand{\xmark}{%
  \tikz[baseline=-0.45ex]\draw[red!72!black,line width=1.05pt,line cap=round]
    (0.08ex,0.08ex)--(0.92ex,0.92ex) (0.08ex,0.92ex)--(0.92ex,0.08ex);%
}

\begin{table}[t]
\centering
\footnotesize
\caption{\textbf{Comparison of humanoid loco-manipulation systems.} \textbf{No Teleop}: no teleoperated demonstrations required. \textbf{Whole-Body}: joint upper- and lower-body control. \textbf{Multi-task}: pipeline reusable across tasks. \textbf{Cost}: cost of scaling up training data for each task. \textbf{--}: no data generation required.}
\label{tab:comparison}
\setlength{\tabcolsep}{2pt}
\renewcommand{\arraystretch}{1.05}
\begin{tabular*}{\columnwidth}{@{\extracolsep{\fill}}lcccc@{}}
\toprule
Method & No Teleop & Whole-Body & Multi-task & Cost\\
\midrule
VIRAL~\citep{he2025viral} & \xmark & \xmark & \xmark & -- \\
DoorMan~\citep{xue2025doorman} & \cmark & \xmark & \xmark & -- \\
HumanoidMimicGen~\citep{lin2026humanoidmimicgen} & \xmark & \xmark & \cmark & Low \\
FetchMan~\citep{rayyan2026fetchman} & \cmark & \xmark & \xmark & Low \\
GRAIL~\citep{xie2026grail} & \cmark & \cmark & \cmark & High \\
\midrule
\textbf{\methodname{} (ours)} & \cmark & \cmark & \cmark & Low \\
\bottomrule
\end{tabular*}
\end{table}

For low-cost reference synthesis, we use GPT-6 Astra~\citep{openai2026astra} to generate reusable task-level interaction templates. These templates specify ordered sequences of key whole-body poses, robot–object and hand–object spatial relations, and intended contact labels. We adapt them to different spatial configurations of the robot, object, and goal to synthesize robot and object trajectories for policy learning.

To reduce task-specific reward design and training effort, we use reference-based imitation as a common learning objective across tasks. We first train privileged teachers to execute these references under physical dynamics, then train visual policies through teacher distillation and reinforcement learning with the same reference-based mimic rewards. Teacher supervision is retained during reinforcement learning to guide policy optimization at states visited by the visual policy. Diverse object instances and visual domain randomization further support visual generalization and sim-to-real transfer.

Across five humanoid loco-manipulation tasks, SUGARv2 achieves a mean success rate of 91.0\%, outperforming the strongest baseline by 51.4 percentage points. Deployed on a real humanoid robot, the visual policies successfully perform whole-body manipulation tasks, demonstrating the framework's potential for learning diverse real-world manipulation skills. Our main contributions are:

\begin{itemize}
\item We present an interaction reference synthesis pipeline for
humanoid loco-manipulation. By adapting reusable task
templates to different spatial configurations, the pipeline
synthesizes diverse interaction references across tasks, reducing the cost of collecting demonstrations.
\item We introduce a policy learning pipeline that uses reference-based imitation as a common learning objective across loco-manipulation tasks. Shared mimic rewards connect privileged teacher training with visual policy learning, reducing task-specific reward design effort.
\item Extensive simulation and real-world experiments validate that our method outperforms strong baselines, generalizes to unseen backgrounds and object instances, and transfers zero-shot to real hardware.
\end{itemize}

\begin{figure*}[t]
\setlength{\abovecaptionskip}{.5em}
\centering
\includegraphics[width=\textwidth]{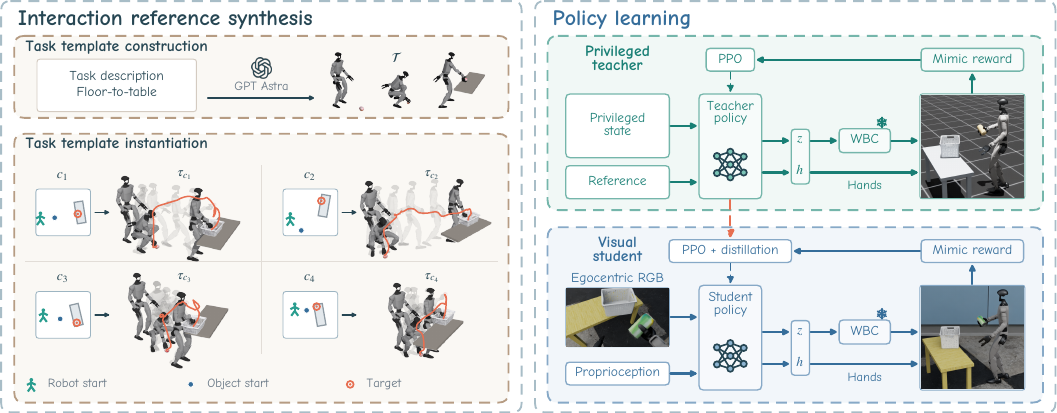}
\caption{Overview of SUGARv2. Left: each task uses an interaction template that is instantiated for different spatial configurations, yielding keyframes and timing to guide continuous interaction reference synthesis. Floor-to-Table is shown as an example, with four configurations $c_1$--$c_4$ and their synthesized references $\tau_{c_1}$--$\tau_{c_4}$. Right: a privileged teacher is trained with a reference-based mimic reward to execute the interaction references through a frozen pretrained whole-body controller (WBC). The visual policy uses the same WBC and learns through teacher distillation and reinforcement learning with the same reward.}
\label{fig:sugarv2}
\end{figure*}

\section{Related Work}
\label{sec:related}

\subsection{Synthetic Demonstrations for Policy Learning}

Demonstration-based methods synthesize interaction references by adapting existing motions~\citep{garrett2025skillmimicgen,jiang2025dexmimicgen,xue2025demogen,fu2025demohlm}. MimicGen~\citep{mandlekar2023mimicgen} adapts object-centric demonstration segments to new configurations. HumanoidMimicGen~\citep{lin2026humanoidmimicgen} extends demonstration adaptation to humanoids using explicitly annotated skills and separate planning for locomotion and stationary manipulation, which constrains whole-body coordination. Generative approaches instead synthesize interaction motions from learned priors~\citep{li2023omomo,li2024chois,diller2024cghoi,jang2025dreamgen,xu2025interact,jung2026dechoi,wang2026prism}. GRAIL~\citep{xie2026grail} generates interaction videos conditioned on 3D assets, reconstructs human and object motion, and retargets the reconstructed motions to humanoid robots.

Beyond generating diverse motions, interaction reference synthesis must account for task requirements and scene geometry. Prior work uses contact relations and scene-adapted keyframes to specify interaction constraints~\citep{yu2026omnicontact,chen2026kino,li2026hoipage}. Conditional motion models generate continuous motion from trajectories, joint constraints, or sparse keyframes~\citep{karunratanakul2023gmd,xie2024omnicontrol,cohan2024condmdi,pinyoanuntapong2025maskcontrol,rempe2026kimodo}. Building on conditional motion generation, SUGARv2 adapts reusable task-level interaction templates to different spatial configurations, uses the adapted key poses to guide whole-body motion generation, and constructs coordinated object trajectories to form interaction references for policy learning.

\subsection{Motion Tracking and Interaction Imitation}

Motion tracking connects kinematic references to physical execution~\citep{peng2018deepmimic,luo2023phc,he2025asap}. Physics-based character controllers can infer motion from sparse conditions~\citep{tessler2024maskedmimic}, while humanoid tracking controllers track motion references and enable whole-body teleoperation~\citep{he2025hover,ze2025twist,liao2025beyondmimic,he2025omnih2o}. Scaling motion data and controller pretraining further broadens the range of motions that can be executed~\citep{luo2026sonic,zeng2026scalebfm}.

Object interaction additionally requires coordinating body motion with object dynamics~\citep{lin2026lessmimic}. The SkillMimic series~\citep{wang2025skillmimic,yu2025skillmimicv2} learns diverse interaction skills through a unified imitation formulation and extends skill coverage from sparse and noisy demonstrations. InterMimic~\citep{xu2025intermimic} uses physics-based teachers to refine imperfect interaction demonstrations and distills their behavior into a general policy. Subsequent work extends this direction through generative control and the augmentation of robot interaction motions~\citep{xu2026interprior,zhang2026intermimicgen}. Reference-based interaction learning also enables humanoid robots to acquire manipulation skills by imitating coordinated human and object motions~\citep{weng2025hdmi,wang2026humanx,wu2026sugar}. These works show how reference-based learning can support interaction skills without a separately designed task reward for each behavior. Building on this, SUGARv2 uses synthesized interaction references to define a common learning objective for visual loco-manipulation across tasks.

\subsection{Visual Humanoid Loco-Manipulation}

Imitation learning trains visual humanoid manipulation policies from demonstrations collected through whole-body teleoperation~\citep{fu2025humanplus,he2025omnih2o}. Humanoid vision-language-action models draw on teleoperated demonstrations, human data, and heterogeneous robot data to learn diverse skills~\citep{nvidia2025grootn1,jiang2025wholebodyvla,hu2026openhlm,xu2026humanverse}. Synthetic demonstrations also support visual policy learning by expanding training data without collecting additional robot demonstrations~\citep{fu2025demohlm,wang2026vlk,kim2026legs}.

Reinforcement learning provides another route to learning visual loco-manipulation skills through interaction with the environment. Some approaches train privileged policies with task-specific rewards and then transfer their behavior to visual policies~\citep{yin2025visualmimic}. VIRAL~\citep{he2025viral} trains visual policies to imitate privileged teachers through online DAgger and behavior cloning. Reinforcement learning can further refine visuomotor behavior~\citep{xue2025doorman,rayyan2026fetchman,he2026ultra}. Extending these approaches to new tasks requires additional reward design and tuning. SUGARv2 uses synthesized interaction references to define a common learning objective across tasks. A shared reference-based mimic reward guides both privileged teacher training and visual policy optimization, reducing task-specific reward design effort.

\section{Method}
\label{sec:method}

\subsection{Problem definition and framework overview}
\label{sec:overview}

We aim to learn visual humanoid loco-manipulation policies that coordinate whole-body motion and object manipulation. Using egocentric images and proprioception, a policy must complete the task across different spatial configurations. We learn a separate policy for each task.

As shown in Figure~\ref{fig:sugarv2}, SUGARv2 shares a two-stage pipeline across tasks: interaction reference synthesis and policy learning. The first stage synthesizes robot motion, object trajectories, and intended contact labels for each spatial configuration (Section~\ref{sec:reference_synthesis}). In the second stage, we train a privileged teacher to execute the synthesized references, then distill its behavior into a visual policy and refine the policy through reinforcement learning with the same reference-based mimic rewards (Section~\ref{sec:policy_learning}).

\begin{figure*}[t]
\setlength{\abovecaptionskip}{.5em}
\centering
\includegraphics[width=\textwidth]{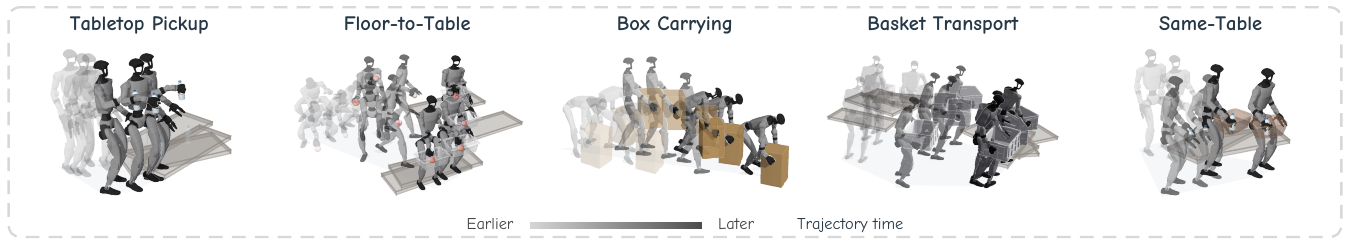}
\caption{Synthesized interaction references across five tasks. Overlaid poses show references for different spatial configurations. Opacity indicates relative time within each displayed reference interval, with later poses shown more opaquely.}
\label{fig:synthesis}
\vspace{-10pt}
\end{figure*}

\subsection{Interaction reference synthesis}
\label{sec:reference_synthesis}

Each task has its own interaction template $\mathcal T$. Given a sampled spatial configuration $c$ specifying the robot's initial state, the object's initial pose, and the goal position, a shared synthesis pipeline $G$ produces an interaction reference $\tau_c=G(\mathcal T,c)$. Each reference is a sequence:
{
\begin{equation}
\tau_c=\left\{\left(q_t^{\mathrm{robot}},q_t^{\mathrm{obj}},l_t\right)\right\}_{t=1}^{T}.
\end{equation}
}
At each time step $t$, the reference contains a whole-body robot pose $q_t^{\mathrm{robot}}$, an object pose $q_t^{\mathrm{obj}}$, and intended contact labels $l_t$ for each hand. $T$ denotes the reference length. These kinematic references are not tied to specific object instances and can be reused with same-category objects during training. Figure~\ref{fig:sugarv2} illustrates how we construct task templates and instantiate them across spatial configurations, and Figure~\ref{fig:synthesis} shows the resulting references.

\subsubsection{Task template construction}

For a given task, motions across spatial configurations typically share the same interaction sequence and contact requirements. For example, carrying an object requires establishing a grasp, transporting the object, and reaching the goal, while the robot's stance, turns, and path depend on the object and goal locations. A task template captures these shared interaction stages, allowing the same task design to be reused across configurations.

For each task, we use GPT-6 Astra~\citep{openai2026astra} to generate an interaction template $\mathcal T$ consisting of an ordered sequence of template keyframes. Each template keyframe specifies a whole-body pose, robot–object and hand–object spatial relations, and intended contact labels. The poses in the template provide starting points for adaptation, while the spatial relations guide robot positioning and hand placement in each configuration. The contact labels specify which hands interact with the object at each template keyframe.

\subsubsection{Task template instantiation}
\label{sec:keyframe_instantiation}

We adapt the template keyframes to each sampled spatial configuration to obtain instantiated keyframes and their timing. Starting from the object's initial pose and the goal position specified by the spatial configuration, we use the template's hand–object relations and the object's symmetries to generate candidate hand targets. We then adapt the robot's position, heading, and whole-body pose to these targets under reachability and scene constraints. Among feasible candidates, we select the keyframe requiring the least movement and pose adjustment. Once the keyframe poses are instantiated, we determine their timing from the movement demands between consecutive poses.

Given the instantiated keyframes and their timing, we use an off-the-shelf motion generation model~\citep{rempe2026kimodo}, additionally conditioned on task text, to generate continuous whole-body motion. For each configuration, we generate multiple candidate trajectories, filter out those that violate kinematic or scene constraints, and rank the remaining candidates to favor smooth motion with less unnecessary movement.

After generating robot motion, we construct the object trajectory from the motion of the interacting hands and the template's hand–object relations. We correct object positions so that the trajectory passes through the positions specified at the keyframes. Using the template's contact labels and keyframe times, we assign contact intervals to the interacting hands. Together, the robot motion, object trajectory, and contact labels form a complete interaction reference for policy learning in the sampled spatial configuration.

\begin{figure*}[t]
\setlength{\abovecaptionskip}{.5em}
\centering
\includegraphics[width=\textwidth]{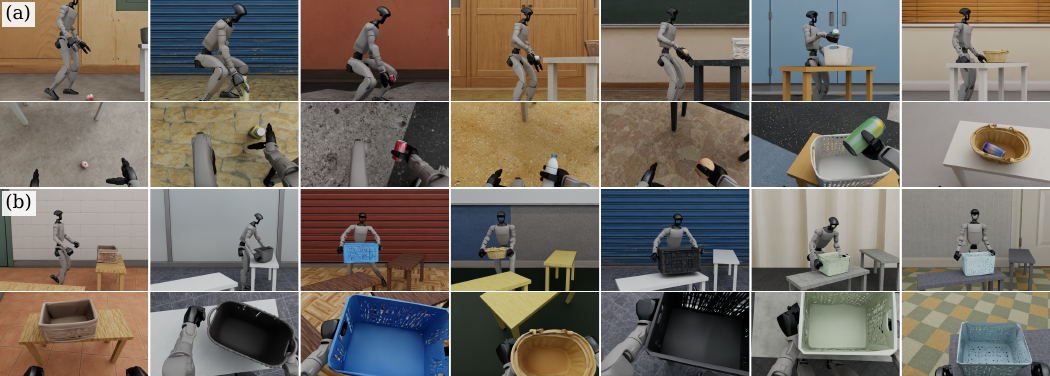}
\caption{Visual diversity across (a) Floor-to-Table and (b) Basket Transport. Each paired example shows the same robot state and scene from a third-person view in the top row and an egocentric RGB view in the bottom row.}
\label{fig:visual_randomization}
\vspace{-10pt}
\end{figure*}

\subsection{Policy learning}
\label{sec:policy_learning}

Building on the synthesized interaction references, our policy learning pipeline uses reference-based imitation as a common learning objective across tasks, reducing task-specific reward design effort. We first train privileged teachers to execute these references, then train visual policies through teacher distillation and reinforcement learning with the same reference-based mimic reward.

\subsubsection{Privileged teacher training}
\label{sec:physical_adaptation}

We train a privileged teacher to execute the interaction references through a frozen pretrained whole-body controller~\citep{luo2026sonic}. The teacher policy $\pi_T(a_t\mid o_t^T)$ receives privileged observations $o_t^T$ comprising privileged states and reference trajectories, and outputs a latent command $z_t$ for the whole-body controller and hand commands $h_t$ specifying whether each hand is open or closed. The whole-body controller executes body motion, while hand commands regulate grasping and release based on reference contact labels.

We train the teacher with a mimic reward comprising tracking, interaction, and regularization terms:
\begin{equation}
r_t=r_t^{\mathrm{track}}+r_t^{\mathrm{int}}+r_t^{\mathrm{reg}}.
\end{equation}
The tracking reward encourages the robot and object to follow their reference motions. The interaction reward encourages their relative poses and hand–object contact states to match the references. The regularization reward penalizes abrupt control changes and undesirable physical behavior. To support visual policy learning, it also encourages the teacher to keep task-relevant objects and goals within the camera's field of view. Together, these terms guide the teacher to reproduce the reference interaction under physical dynamics. This reference-based reward formulation supports a common training procedure across tasks.

Training episodes start from reference states before the first intended contact, with random perturbations applied to the robot and object states. We apply early termination when the robot or object deviates substantially from the reference or enters an invalid physical state. Reward definitions and training details are provided in the appendix.

\subsubsection{Visual policy learning}
\label{sec:visual_learning}

We train visual policies to perform manipulation from egocentric images, using privileged teachers for action supervision. The policy $\pi_S(a_t\mid o_t^S)$ receives egocentric RGB images and proprioceptive inputs $o_t^S$. A visual encoder~\citep{qin2024mobilenetv4} projects image features and concatenates them with proprioception, and an MLP predicts the commands $(z_t,h_t)$.

Learning directly from images requires jointly learning visual representations and control. We first train the visual policy through teacher distillation, querying teacher actions at states visited by the policy~\citep{ross2011dagger}. Matching individual teacher actions does not directly optimize the full interaction, so we continue with the reference-based mimic reward while retaining teacher distillation, optimizing robot and object motion throughout the interaction.

The frozen teacher provides the complete body-control latent and hand commands as action supervision at student-visited states. We minimize the mean squared error between the student’s predicted action and the teacher target:
\begin{equation}
\mathcal L_{\mathrm{distill}}=\operatorname{MSE}\!\left(a_t^{S},a_t^{T}\right),
\end{equation}
where $a_t^{S}$ is the student’s predicted action, $a_t^{T}$ is the teacher target at the visited state.
We jointly optimize the visual policy with PPO~\citep{schulman2017ppo} and teacher distillation:

\begin{equation}
\mathcal L_{\mathrm{policy}}
=\mathcal L_{\mathrm{PPO}}^{\mathrm{actor}}
+\lambda_{\mathrm{distill}}\mathcal L_{\mathrm{distill}},
\end{equation}

where $\mathcal L_{\mathrm{PPO}}^{\mathrm{actor}}$ is the PPO policy loss and $\lambda_{\mathrm{distill}}$ weights action distillation during policy optimization.

The entire visual policy is trained end-to-end throughout training. A privileged critic is warmed up during distillation and updated alongside the policy during joint optimization.

We render observations online with diverse object instances, backgrounds, materials, and lighting conditions. Each interaction reference can provide training data under varied visual appearances without regenerating the underlying motion. Figure~\ref{fig:visual_randomization} illustrates this visual diversity.

During visual training, we randomize camera poses, augment rendered images, and add noise to images and proprioceptive observations. We also use dynamics randomization and external disturbances in both training stages to improve robustness to physical variations.

\section{Experiments}
\label{sec:experiments}

Through our experiments, we aim to answer the following questions: (1) How well does SUGARv2 learn visual manipulation skills under varied spatial configurations and unseen backgrounds? (2) How efficiently can we synthesize interaction references, and how well do they support visual policy learning? (3) How does the number of synthesized references affect visual policy performance? (4) How do key design choices in our learning framework affect policy performance? (5) Can the learned visual policies transfer to a real humanoid robot?

\subsection{Experimental Setup}

\noindent\textbf{Tasks.} We evaluate five humanoid loco-manipulation tasks covering pickup, pick-and-place, and object transport. \textbf{\textit{Tabletop Pickup}} requires grasping an object from a tabletop and holding it. \textbf{\textit{Floor-to-Table}} and \textbf{\textit{Same-Table}} involve picking up an object from the floor or tabletop, respectively, and placing it into a target basket on a table. \textbf{\textit{Box Carrying}} and \textbf{\textit{Basket Transport}} require carrying the corresponding object while walking to a designated target region. These tasks demand tightly coordinated whole-body control, requiring continuous coordination between the upper and lower body throughout object interaction.

\begin{table*}[t]
\centering
\small
\caption{Visual policy success rates (\%) across five tasks. (a) VIRAL and GR00T using SUGARv2 references and teachers. (b) Our learning pipeline using GRAIL references. (c) The full SUGARv2 framework combining our reference synthesis and visual policy learning pipeline.}
\label{tab:main}
\setlength{\tabcolsep}{3pt}
\renewcommand{\arraystretch}{1.05}
\begin{tabular*}{\textwidth}{@{\extracolsep{\fill}}cccccccc@{}}
\toprule
Data source & Learning method & Tabletop Pickup & Floor-to-Table & Box Carrying & Basket Transport & Same-Table & Mean\\
\midrule
\multicolumn{8}{@{}l}{\textit{(a) Comparing learning methods}}\\
SUGARv2 & VIRAL & $63.2\!\pm\!0.9$ & $13.4\!\pm\!4.6$ & $20.4\!\pm\!9.1$ & $41.4\!\pm\!1.2$ & $59.7\!\pm\!4.8$ & 39.6\\
SUGARv2 & GR00T N1.7 & $38.9\!\pm\!2.3$ & $13.8\!\pm\!2.4$ & $15.4\!\pm\!1.4$ & $20.6\!\pm\!3.4$ & $28.2\!\pm\!2.2$ & 23.4\\
\midrule
\multicolumn{8}{@{}l}{\textit{(b) Comparing interaction reference sources}}\\
GRAIL & SUGARv2 & $43.3\!\pm\!0.7$ & $0.6\!\pm\!0.2$ & $83.1\!\pm\!1.3$ & $11.7\!\pm\!2.1$ & $2.1\!\pm\!0.1$ & 28.2\\
\midrule
\multicolumn{8}{@{}l}{\textit{(c) Full SUGARv2 framework}}\\
\textbf{SUGARv2} & \textbf{SUGARv2} & $\mathbf{92.8\!\pm\!1.5}$ & $\mathbf{91.7\!\pm\!3.0}$ & $\mathbf{87.6\!\pm\!1.9}$ & $\mathbf{84.3\!\pm\!2.3}$ & $\mathbf{98.5\!\pm\!0.4}$ & \textbf{91.0}\\
\bottomrule
\end{tabular*}
\end{table*}

\noindent\textbf{Training and evaluation.} We train visual policies across diverse spatial configurations of robot, object, and goal within prescribed ranges. We use 100 synthesized references per task and train separate privileged teachers and visual policies with the same pipeline. In simulation, we report full-task success as mean $\pm$ standard deviation over three seeds, using task-specific criteria.

\noindent\textbf{Baselines.} For visual policy learning, we compare with \textbf{\textit{VIRAL}}~\citep{he2025viral} and \textbf{\textit{GR00T N1.7}}~\citep{nvidia2025grootn1}. VIRAL combines DAgger with behavior cloning. GR00T is fine-tuned on image–action pairs from teacher rollouts. For interaction reference synthesis, we compare with \textbf{\textit{GRAIL}}~\citep{xie2026grail}, which reconstructs motion from generated videos.

\subsection{Visual Policy Performance}

\noindent\textbf{Comparison with baselines.} We compare visual policy learning methods using the same interaction references and privileged teachers. Table~\ref{tab:main}(a) and (c) report full-task success rates across five tasks. SUGARv2 outperforms the baseline methods on all five tasks, achieving a mean success rate of 91.0\%. Under the same references and teachers, the consistent improvements across tasks demonstrate the effectiveness of our learning pipeline in translating synthesized references into diverse visual manipulation skills.

\begin{table}[t]
\centering
\small
\caption{SUGARv2 success rates (\%) under unseen backgrounds and object instances, without additional training.}
\label{tab:visual_generalization}
\setlength{\tabcolsep}{0.5pt}
\renewcommand{\arraystretch}{1.0}
\begin{tabular*}{\columnwidth}{@{\extracolsep{\fill}}lcccc@{}}
\toprule
Task & Original & \shortstack{Unseen\\backgrounds} & \shortstack{Unseen\\objects} & \shortstack{Both\\unseen}\\
\midrule
Tabletop Pickup & $\mathbf{92.8\!\pm\!1.5}$ & $90.6\pm0.3$ & $86.9\pm1.1$ & $83.7\pm1.0$\\
Floor-to-Table & $\mathbf{91.7\!\pm\!3.0}$ & $85.3\pm1.9$ & $89.9\pm0.7$ & $80.0\pm2.2$\\
Box Carrying & $\mathbf{87.6\!\pm\!1.9}$ & $83.5\pm1.4$ & $85.2\pm1.4$ & $81.6\pm1.7$\\
Basket Transport & $84.3\!\pm\!2.3$ & $77.8\pm1.6$ & $\mathbf{88.2\pm0.7}$ & $83.6\pm0.4$\\
Same-Table & $\mathbf{98.5\!\pm\!0.4}$ & $97.2\pm0.1$ & $88.0\pm0.4$ & $87.1\pm0.3$\\
\midrule
Mean & \textbf{91.0} & 86.9 & 87.7 & 83.2\\
\bottomrule
\end{tabular*}
\end{table}

\noindent\textbf{Generalization to unseen backgrounds and objects.}
We evaluate SUGARv2 on unseen backgrounds and same-category object instances without further training, changing backgrounds and objects separately and jointly while keeping spatial configurations and success criteria consistent. Table~\ref{tab:visual_generalization} compares success rates with the original setting. Success rates decrease under these conditions but remain high overall, indicating generalization to unseen backgrounds and same-category object instances.

\begin{table*}[t]
\centering
\small
\caption{Task success rates (\%) for ablations of the learning strategy.}
\label{tab:abl_design}
\setlength{\tabcolsep}{3pt}
\renewcommand{\arraystretch}{1.15}
\begin{tabular*}{\textwidth}{@{\extracolsep{\fill}}lcccccc@{}}
\toprule
Setting & Tabletop Pickup & Floor-to-Table & Box Carrying & Basket Transport & Same-Table & Mean\\
\midrule
\textbf{Full SUGARv2} & $\mathbf{92.8\!\pm\!1.5}$ & $\mathbf{91.7\!\pm\!3.0}$ & $\mathbf{87.6\!\pm\!1.9}$ & $\mathbf{84.3\!\pm\!2.3}$ & $\mathbf{98.5\!\pm\!0.4}$ & \textbf{91.0}\\
\midrule
w/o teacher distillation & $0.1\!\pm\!0.1$ & $0.0\!\pm\!0.0$ & $0.0\!\pm\!0.0$ & $0.0\!\pm\!0.0$ & $0.0\!\pm\!0.0$ & 0.0\\
w/o reference-based RL & $80.5\!\pm\!1.8$ & $77.1\!\pm\!6.5$ & $3.6\!\pm\!2.1$ & $3.9\!\pm\!2.6$ & $42.0\!\pm\!25.3$ & 41.4\\
w/o continued teacher distillation & $79.0\!\pm\!2.5$ & $61.6\!\pm\!3.3$ & $60.9\!\pm\!7.9$ & $75.2\!\pm\!5.2$ & $76.6\!\pm\!4.1$ & 70.6\\
\bottomrule
\end{tabular*}
\end{table*}

\begin{table}[t]
\centering
\small
\caption{Reference synthesis cost and teacher performance, averaged across five tasks. Synthesis time includes failed attempts and postprocessing per retained reference. Tracking errors measure deviations of executed motions from references.}
\label{tab:synth}
\setlength{\tabcolsep}{4pt}
\renewcommand{\arraystretch}{1.12}
\begin{tabularx}{\columnwidth}{@{}>{\raggedright\arraybackslash}p{0.52\columnwidth}*{2}{>{\centering\arraybackslash}X}@{}}
\toprule
Metric & GRAIL & SUGARv2\\
\midrule
Synthesis time (min/ref.) $\downarrow$ & 36.1 & \textbf{0.4}\\
Teacher success (\%) $\uparrow$ & 62.4 & \textbf{98.0}\\
Object tracking error (cm) $\downarrow$ & 9.8 & \textbf{6.6}\\
Robot tracking error (cm) $\downarrow$ & 13.8 & \textbf{10.0}\\
\bottomrule
\end{tabularx}
\end{table}

\subsection{Reference Synthesis Evaluation}

We measure synthesis cost and assess reference quality through physical execution and downstream visual policy performance. SUGARv2 and GRAIL generate the same number of references per task for the same desired spatial configurations. For each source, we train separate privileged teachers and visual policies with the same learning pipeline and training budget.

As shown in Table~\ref{tab:synth}, SUGARv2 takes an average of 0.4 minutes per retained reference, compared with 36.1 minutes for GRAIL. The resulting references yield higher teacher success and lower tracking errors, indicating that they better support task execution under physical dynamics. GRAIL can introduce physically implausible object motion during video generation and depth or pose errors during reconstruction. SUGARv2 explicitly constrains robot–object spatial relations and derives object trajectories from hand motion, helping preserve interaction consistency.

Downstream visual policy performance further supports this assessment of reference quality. With the learning pipeline and budget held fixed, policies trained on SUGARv2 references outperform those trained on GRAIL references across all five tasks (Table~\ref{tab:main}(b) and (c)). This consistent advantage supports the higher quality of our synthesized references for visual policy learning.

\subsection{Influence of Reference Data Size}

We examine how reference data scale affects visual policy learning. We use 20, 50, 100, 200, and 500 references per task while keeping the training pipeline and budget fixed. Data scale counts kinematic robot–object trajectories with contact labels across spatial configurations. During training, we vary object instances within each category and render visual observations from the simulation, using the same visual randomization settings across experiments.

\begin{figure}[t]
\centering
\includegraphics[width=\columnwidth]{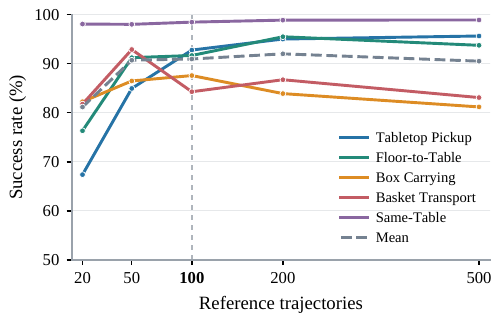}
\caption{Visual policy success rates at different reference data scales. The dashed line denotes the mean across five tasks. Horizontal positions follow $x^{0.6}$; tick labels show trajectory counts.}
\label{fig:scale}
\end{figure}

Figure~\ref{fig:scale} shows that increasing the reference set from 20 to 100 trajectories per task generally improves policy success rates, while further increases offer limited gains and sometimes reduce performance. We therefore use 100 references per task in the main evaluation, balancing policy performance with the cost of reference synthesis.

\subsection{Ablation study}
We evaluate the key design choices in our visual policy learning pipeline through ablations under the same training budget (Table~\ref{tab:abl_design}). Without teacher distillation, training the policy from scratch using reference-based reinforcement learning yields near-zero success on all five tasks under this training budget. Without reference-based reinforcement learning, training through teacher distillation alone results in lower success than the full framework on every task. Finally, after the same initialization through teacher distillation, continuing with reference-based reinforcement learning alone reduces success across all tasks compared with retaining teacher distillation.

These results support the design of our visual policy learning pipeline: teacher distillation provides an effective initialization, while reference-based reinforcement learning refines task performance with continued teacher supervision. Together, these components enable effective learning of visual loco-manipulation skills.

\begin{figure*}[t]
\centering
\includegraphics[width=\textwidth]{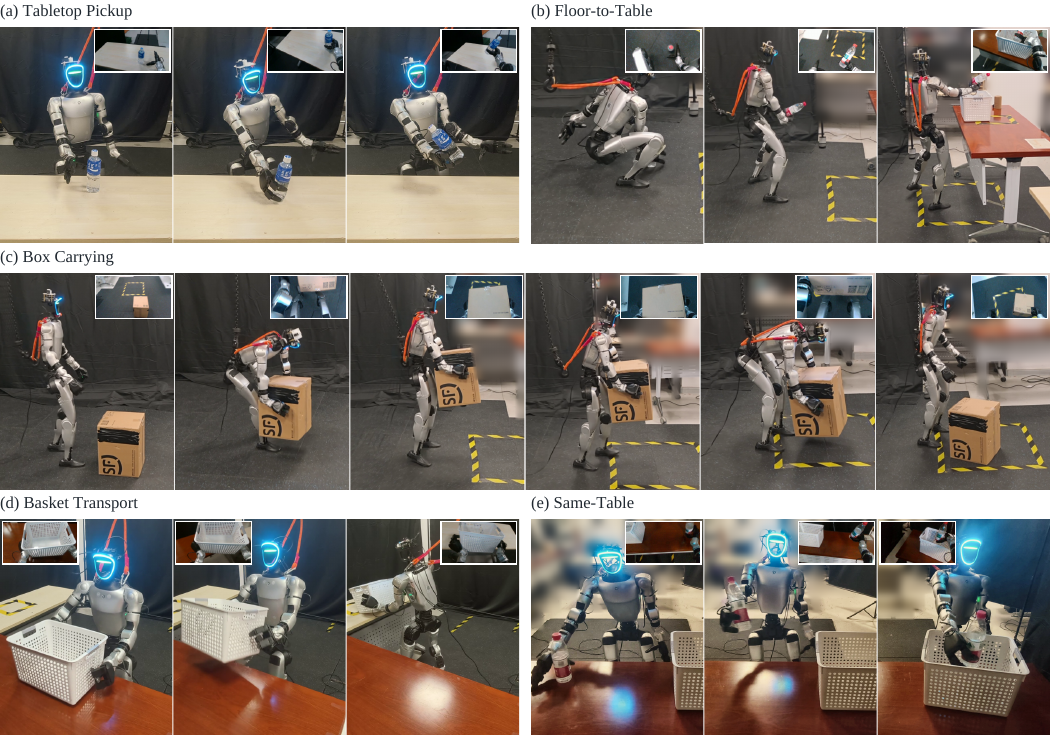}
\caption{Real-world execution: (a) Tabletop Pickup, (b) Floor-to-Table, (c) Box Carrying, (d) Basket Transport, (e) Same-Table.}
\label{fig:real}
\end{figure*}

\begin{table}[t]
\centering
\small
\caption{Real-world task success (successes / attempts).}
\label{tab:real}
\setlength{\tabcolsep}{2pt}
\renewcommand{\arraystretch}{1.12}
\begin{tabularx}{\columnwidth}{@{}*{5}{>{\centering\arraybackslash}X}@{}}
\toprule
\shortstack{Tabletop\\Pickup} & \shortstack{Floor-to-\\Table} & \shortstack{Box\\Carrying} & \shortstack{Basket\\Transport} & \shortstack{Same-\\Table}\\
\midrule
9/10 & 6/10 & 8/10 & 8/10 & 7/10\\
\bottomrule
\end{tabularx}
\vspace{-5pt}
\end{table}

\subsection{Real-world evaluation}

\noindent\textbf{Real-world transfer.} We directly deploy the simulation-trained visual policies on a Unitree G1 without fine-tuning on any real-world data. Across trials, we vary the initial positions. Table~\ref{tab:real} reports successful trials out of total attempts for each task, and Figure~\ref{fig:real} presents representative executions. The policies successfully perform all five tasks on the real robot, demonstrating sim-to-real transfer for coordinated whole-body motion and object manipulation.

\noindent\textbf{Failure analysis.} Execution errors can cause missed grasps, after which the policy may continue subsequent motions without securing the object. Recovery from these failures remains limited, as the policy does not consistently retry the grasp before proceeding.

\section{Conclusion}
In this work, we presented SUGARv2, a unified framework for interaction reference synthesis and visual policy learning across tasks. We adapt reusable task templates to different spatial configurations to obtain diverse interaction references at low cost. These references support a unified training pipeline combining teacher distillation and reinforcement learning with shared reference-based mimic rewards, reducing task-specific reward design effort. Experiments across five tasks and real-humanoid deployment demonstrate the framework's effectiveness, highlighting its potential for learning diverse real-world manipulation skills.

\vspace{7.59pt}
\noindent\textbf{Limitations and Future Works.} Our reference synthesis currently focuses on short-horizon interactions in simplified scenes. Extending it to long-horizon tasks and cluttered environments will require more scalable and automated interaction reference synthesis. Our visual policies are task-specific; we aim to use synthesized references to support multitask vision-language-action policies that follow language instructions across diverse interaction tasks.

\clearpage
\section*{LLM Usage Statement}

{\looseness=-1 We used GPT-6 Astra~\citep{openai2026astra} to generate reusable task-level interaction templates. We used OpenAI GPT models and Codex to assist with language editing and code development, and reviewed the edited text.\par}

\hypersetup{allcolors=cvprblue}
{\small
\bibliographystyle{ieeenat_fullname}
\bibliography{references}
}

\end{document}